# Plant Species Classification Using Transfer Learning by Pre-trained Classifier VGG-19

Thiru Siddharth, Bhupendra Singh Kirar, and Dheeraj Kumar Agrawal

Thiru Siddharth is with the Department of Computer Science and Engineering, Indian Institute of Information Technology, Bhopal, 462003, India (e-mail: thirusid789@gmail.com).

Bhupendra Singh Kirar is with Department of Electronics and Communication Engineering, Indian Institute of Information Technology, Bhopal, 462003, India (corresponding author phone: 9893411079; e-mail: bhup17@gmail.com).

Dheeraj Kumar Agrawal is with Department of Electronics and Communication Engineering, Maulana Azad National Institute of Technology, Bhopal, 462003, India (e-mail: profdheerajagrawal@gmail.com).

**ABSTRACT**

Deep learning is currently the most important branch of machine learning, with applications in speech recognition, computer vision, image classification, and medical imaging analysis. Plant recognition is one of the areas where image classification can be used to identify plant species through their leaves. Botanists devote a significant amount of time to recognizing plant species by personally inspecting. This paper describes a method for dissecting color images of Swedish leaves and identifying plant species. To achieve higher accuracy, the task is completed using transfer learning with the help of pre-trained classifier VGG-19. The four primary processes of classification are image preprocessing, image augmentation, feature extraction, and recognition, which are performed as part of the overall model evaluation. The VGG-19 classifier grasps the characteristics of leaves by employing pre-defined hidden layers such as convolutional layers, max pooling layers, and fully connected layers, and finally uses the soft-max layer to generate a feature representation for all plant classes. The model obtains knowledge connected to aspects of the Swedish leaf dataset, which contains fifteen tree classes, and aids in predicting the proper class of an unknown plant with an accuracy of 99.70% which is higher than previous research works reported.

## 1. INTRODUCTION

Plants are very essential for life, as they are very valuable source of oxygen and food for all living creatures on earth. Moreover, plants help to clean water and also provide us many valuable ingredients which are used to make many types of medicines. Plants use chlorophyll to turn sunlight and carbon dioxide into oxygen and sugar (helps plants grow). On this planet there are 321000+ plant species. For botanists it is very hard and time-consuming process to study about all plant species. Because of this reason, a plant species classification system is required for the stability of biodiversity, and it can also help an individual for general uses [1,2].

Plant categorization is a difficult undertaking that mainly entails observing morphological traits (like the global characteristics, root, stem, leaf structure, and fruit). Plant leaves include a lot of information concerning plant species recognition. Furthermore, the life span of a leaf is substantially longer than that of other features (like fruit and flowers). As a result, leaf image collections are used in a significant number of plant species identification models. Many recent suggested research projects on plant species identification have shown positive results, but more precise models are still needed.

Previously proposed classification models on leaf prediction are mostly done by the CNN architecture in which all CNN layers needed to be defined from scratch. Despite this they are lacking in accuracy. Therefore, an approach that overcome all the loopholes of recently published work need to be proposed.

This paper presents a more easy and convenient approach to plant species recognition using transfer learning by pre-trained classifier VGGNet. We have experimented model on Swedish leaf dataset. The experimental achieved results show that the proposed model outperforms other plant leaf recognition approaches. This method is simple to adopt and can considerably increase plant species recognition accuracy. This approach can help researcher to identify plant species and can save the inestimable time of botanists. This method of plant recognition can also assist ordinary people who have no prior understanding of plants in identifying and using plants for their purposes.

The remaining part of the paper is written in the following manner: In Section 2, we take a look at some recent research. Section 3 explains the proposed methodology. The identification results that we have achieved are described in Section 4. In Sections 5 and 6, there is a discussion and a conclusion, respectively.

## 2. EXISTING WORK

A review of the existing work for plant leaf is described in this section.

Jeon and Rhee [3] proposed a GoogleNet model for plant categorization by utilizing the CNN architecture. The performance of each model is then assessed by taking leaf damage into account. Lee et al. [4] suggested a two-stream CNN model for plants classification. They used CNN and a deconvolutional network (DN) technique to obtain intuition of the chosen features, and then merged both to classify plant species. Shah et al. [5] presented a dual-path DCNN to train joint visual features for leaf images, based on their morphology and texture properties, and to optimize these features for classification. Hu et al. [6] developed a Multi-Scale-Fusion (MSF)-CNN model for plant leaf identification at multiple scales. In this method, a set of bilinear interpolation procedures is used to down sample an input picture into numerous low-resolution images. The MSF-CNN architecture then learns to discriminate between different depths by gradually feeding these input pictures of varied sizes into it. Multiscale pictures are gradually handled and the associated features are merged as the depth of the MSF-CNN increases. Turkoglu and Hanbay [7] suggested different LBP-based techniques for recognizing plant leaves by utilizing extracted textural information from plant leaves. The suggested approaches use the red and green color channels of pictures, whereas the original LBP reduces color images to grey tones. They also test the suggested approaches' resilience against noise, such as Gaussian and salt and pepper. The extreme learning machine (ELM) approach was used to classify and test the acquired features from the presented methodologies. Manasa et al. [8] classed plants according to their leaf structure. Preprocessing was done of the acquired image, which included scaling, image enhancement, shadow reduction, and background removal. When numerous leaves encircled a leaf, the watershed method was utilized to separate each leaf for image segmentation. Using a neural network, this approach can properly categorize eight distinct plant species. The method's ultimate findings are tailored, concise, and instructive. On a Swedish leaf dataset, the model had an accuracy of 80%. Zhu et al. [9] suggested a new strategy leveraging a deep convolutional neural network called the two-way attention model. As the name implies, it has two modes of attention: the first mode is family first attention, which is based on traditional plant taxonomy and seeks to recognize the plant's family. The second method of attention is max-sum attention, which finds the max-sum section of the fully convolutional network heat map and concentrates on the discriminative properties of the input picture. This approach reduces family label conflicts and improves classification accuracy. Chengzhuan [10] introduced a better technique for classifying plant leaves that incorporates shape and texture features, and texture feature being used as the local binary pattern histogram Fourier (LBP-HF). On the Swedish leaf dataset, with 1-NN approach attained an accuracy of 98.48%. Sumedh et al. [11] attempted a comparative examination of several plant identification methods. They demonstrated a number of experiments using Swedish leaves, including VGG-16, which achieved 95.75 percent accuracy on the Swedish leaf dataset, demonstrating the effectiveness of machine learning and CNN-based classification models. Dhananjay [12] proposed a mechanism for classification of plant from its leaves using CNN particularly for Swedish leaves. proposed work has achieved the good accuracy for Swedish leaf dataset.

Compared with above methods, proposed method achieves comparable performance for leaf recognition with high interpretability.

## 3. PROPOSED METHODOLOGY

In our work, we have leveraged an efficient pre-trained classifier VGG19 based on DNN architecture for recognizing plant species with the help of leaf images. The proposed model has four steps: image preprocessing, image augmentation, feature extraction and model evaluation. Model is presented in Figure 1.

### 3.1 Dataset

The VGG-19 model was trained using the Swedish leaf dataset in this research [13]. Dataset was designed and developed by Söderkvist [14], Kumar et al. [15], Kumar and Rao [16] from computer vision laboratory. This dataset is used to evaluate plant species because of its appealing properties. It's a widely used dataset. The dataset consists of 15 plant classes, each with approximately 75 photos, for a total of approximately 1125 images. The suggested model is trained, validated, and tested using photos from this dataset. Figure 2 depicts the plant species in each of the 15 groups.

### 3.2 Image Pre-processing

Before feeding the images into the model, image preprocessing is used to improve the image data. Rotation, resizing, normalizing, rescaling (grey scaling), and shearing are all examples of preprocessing [13,17]. Preprocessing is usually done in one of two situations: (a). Data Cleaning (b). Data Augmentation.

#### 3.2.1 Data Cleaning

Data cleaning is used to describe the process of removing artefacts from data using a variety of transformations in order to improve the learning process of a model.

#### 3.2.2 Image Augmentation

Image augmentation creates artificial training pictures by utilizing a range of processing techniques such as random rotation, shifting, shearing, flipping, and zooming, among others. To put it another way, image augmentation is a data augmentation approach that produces more training data images from existing training samples.

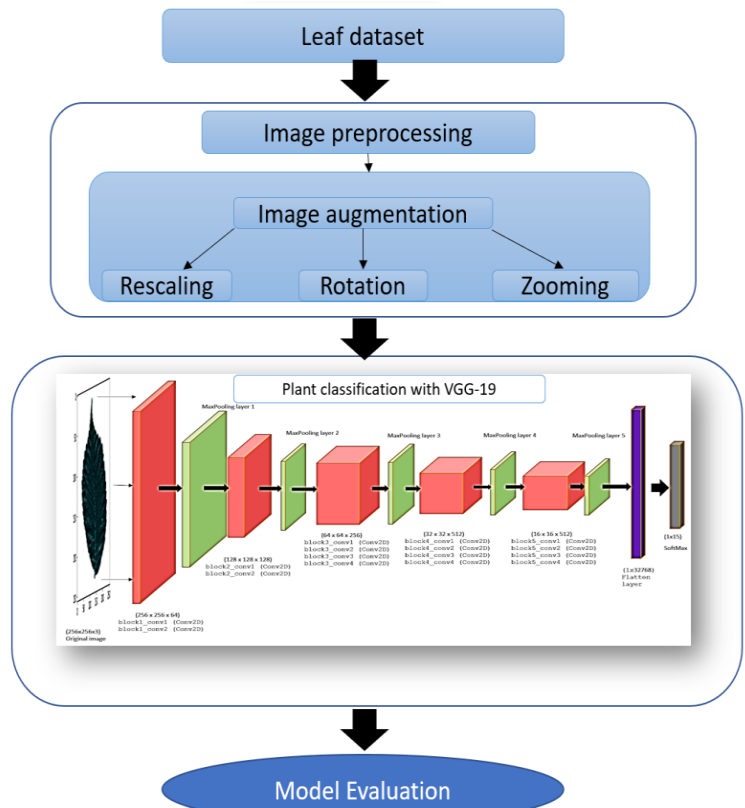

**Figure 1.  Block diagram of the process of plant species classification using VGG-19.**

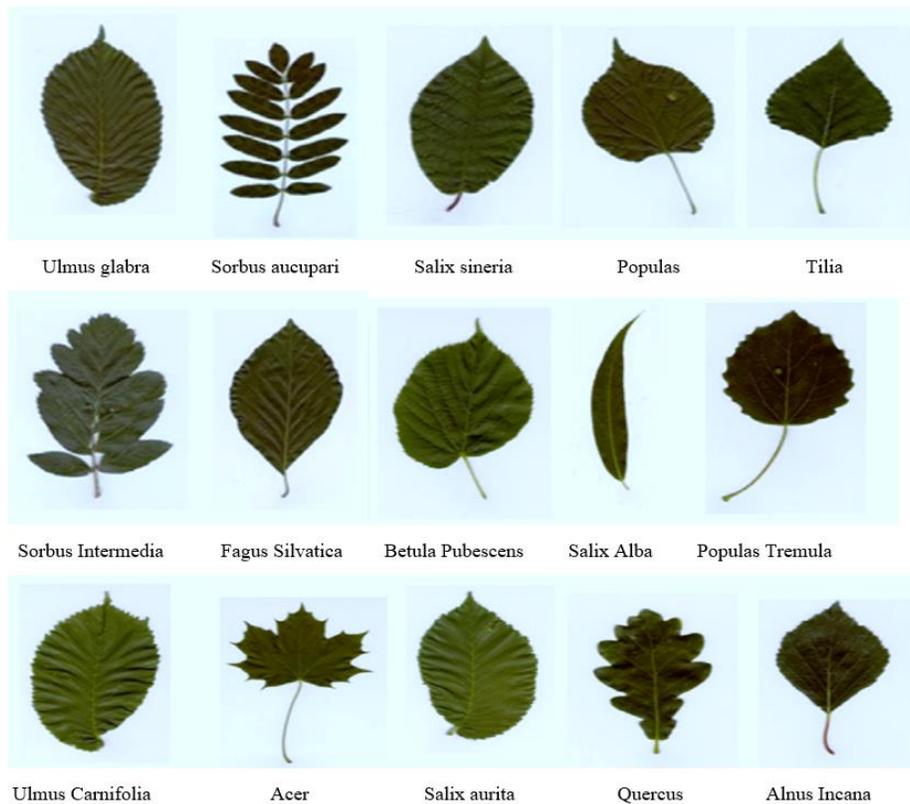

**Figure 2. Images of the Swedish leaves, one image per class.**

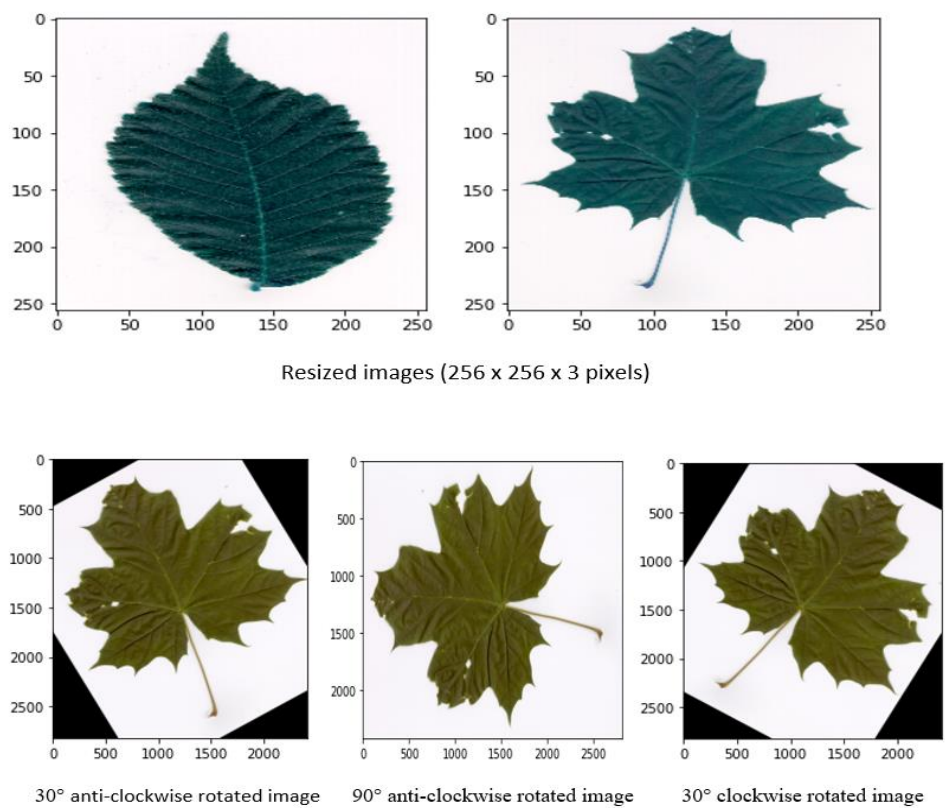

Figure 3. Pre-processed and augmented leaf's images.

Table 1: Image augmentation

| S. No. | Operation | Properties |
|---|---|---|
| 1 | Resize image | 256×256×3 pixels |
| 2 | Rescale | To normalize the input, 1/255 was multiplied by the image channel values. |
| 3 | Rotation range | 30 ° clockwise and anti-clockwise |
| 4 | Zoom range | 0.1% smaller or larger |
| 5 | Validation split | 30% for validation data 70% for training data |

In Data cleaning and augmentation were performed in the suggested system, with various transformation operations applied to the existing samples and more training images generated. Table 1 and Figure 3 show the results.

### 3.3  VGG-19 (VGGNet)

The VGG19 belongs to visual geometry group (VGG) model, which has 19 layers in total (SoftMax=1 layer, MaxPool layers=5, fully connected layers=3, and convolution layers=16). Other VGG variations include VGG11, VGG16, and more. There are 19.6 billion FLOPs in VGG19. The VGG-19 model becomes simpler and more useful as 3 x 3 convolutional layers are layered on top of each other to increase with depth level [18,19]. In VGG-19, Max pooling layers were utilized as a handler to minimize the volume size. Only one FC layer was applied. The resized images were used as input data for the VGGNet DNN, as seen in Figure 4.

The suggested model accepts a 256 x 256 x 3 RGB picture as input. In CNN, four layers, namely, convolution layer,

reLU layer, pooling layer, and fully connected layer are used to extract information from an image. A convolution procedure is performed by many feature filters in the convolution layer. These features will compare two small portions of larger photos and classify the image appropriately if they match. This layer will require four steps: first, it will need to line up the feature filter in the image, and then it will need to multiply pixel of each image by the associated feature pixel. After that, add the values and divide them by the total number of pixels in the feature to get the sum. The filtered image's final observed value is displayed in the middle. Similarly, the feature filter sweeps all around the image and repeats the same steps as before. This technique is performed for each feature filter to obtain the convolution output. ReLU is another type of rectified linear activation function, which outputs the input directly if it is positive and 0 otherwise, as shown in Figure 5(c). It is a default activation function for many types of neural networks due to faster and better performer. It is used to create the output by applying it to all of the rectified feature map (feature images).

Sigmoid is also another sort of activation function, also known as a squashing function as shown in Figure 5(a) as an S-shape. The outputs of the sigmoid function are contained within the 0 and 1 limits. It shrinks the pictures in the pooling layer. After passing through the activation layer, it applies to filtered images. It includes the logistic function, the hyperbolic tangent (Tanh) Figure 5(b), and the arctangent. In the process of pooling, the window size (2 x 2 or 3 x 3) will be selected initially, and the window will traverse across the filtered image, extracting the maximum value from each window.

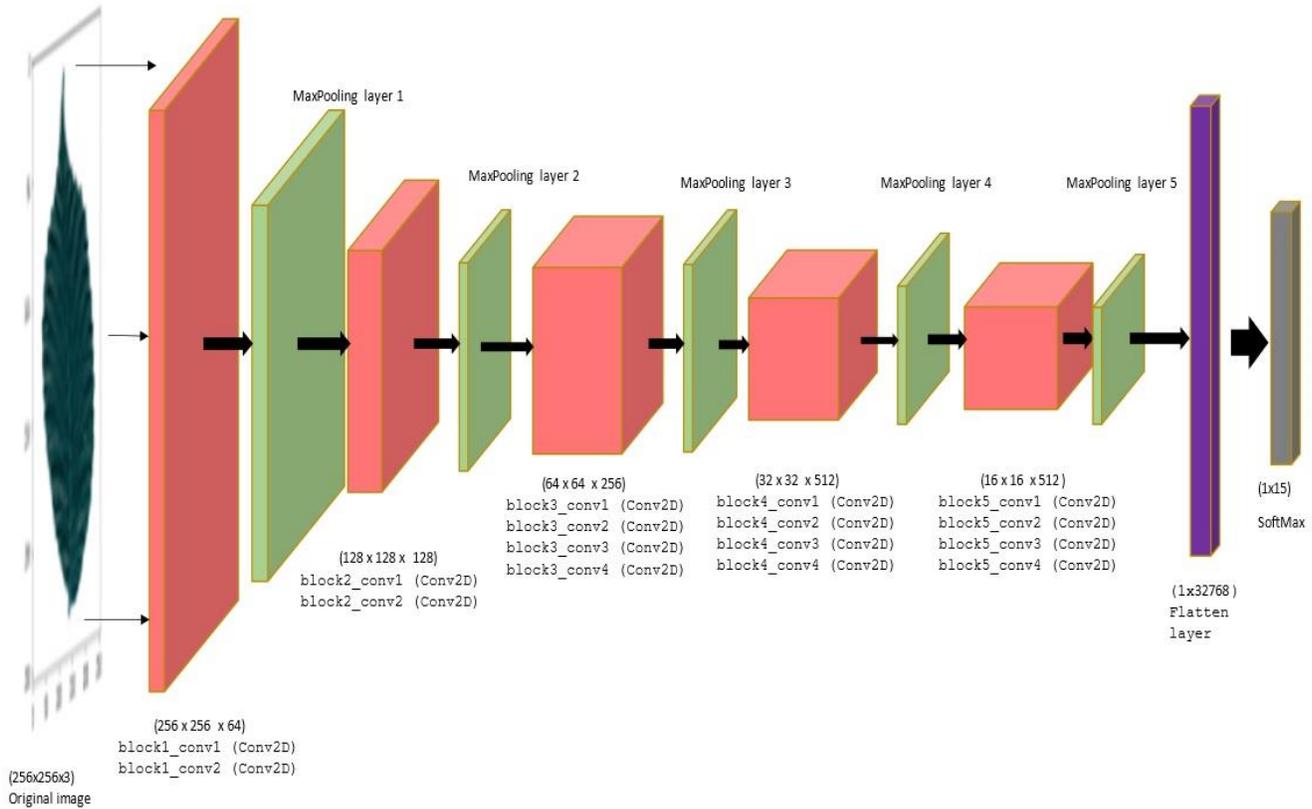

**Figure 4. Leaf classification process with architecture of VGG19 model.**

Finally, the pooled feature map is flattened by the flattening layer and then sent to the fully connected or dense layer, which does the actual classification and predicts the end output using the softmax activation function. Softmax is a type of probabilistic logistic regression. It creates a probability distribution from a set of values. The output vector's elements are in the range of (0, 1) and add up to 1. Given using

Equation (1). It is also known as maximum entropy classifier.

$$p\left(y=j\middle|\theta^{(i)}\right)=\frac{e^{\theta^{(i)}}}{\sum_{j=0}^{k}e^{\theta_k^{(i)}}}\ .Where\ \theta=w_0x_0+w_1x_1+\dots+w_kx_k=\sum_{i=0}^{k}w_ix_i=w^Tx$$

(1)

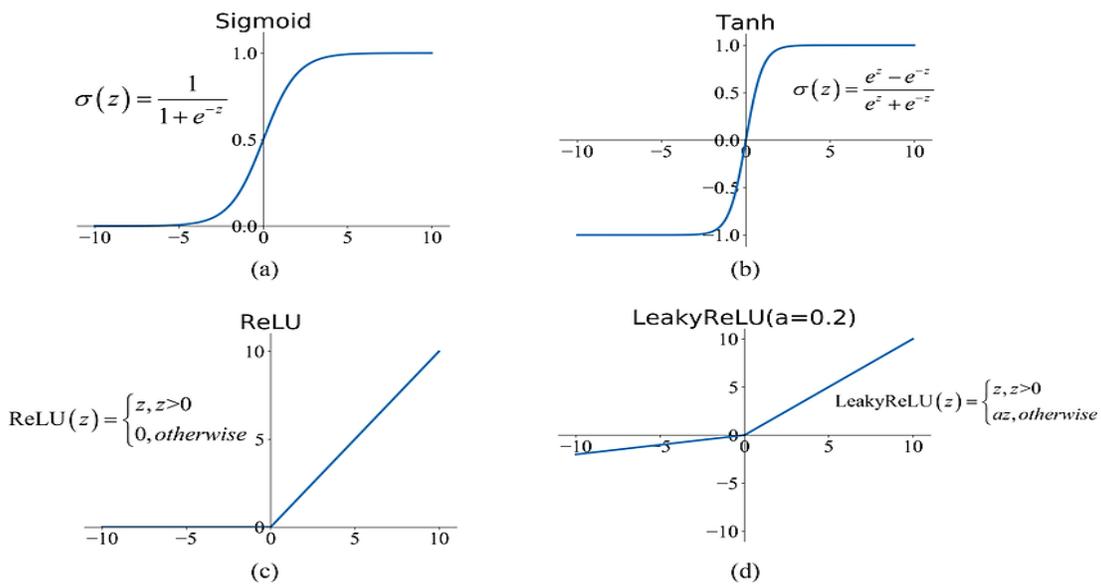

**Figure 5. Activation functions of CNN (a) sigmoid (b) Tanh (c) ReLU (d) LeakyReLU.**

### 3.3.1 Training

The computer first scans the images, which are made up of three fundamental colors called RGB (Red, Green, and Blue). The pixel values for each color are different. For example, if a picture has the dimensions A x B x 3, it indicates it has A rows, B columns, and three colors. Consider the image 48 x 48 x 3, which implies 48 rows, 48 columns, and 3 colors. This is how the computer interprets the image. Convolutional layers were utilized for feature extraction in the training phase, and max pooling layers were combined with some of the convolutional layers to reduce feature dimensionality. In total, 64 kernels (3 x 3 filter size) were used in the first convolutional layer to extract features from the input pictures. The feature vector was created using a fully connected layer or a dense layer.

### 3.3.2 Testing

Finally, throughout the testing phase, we kept the batch size constant at 50 and calculated validation accuracy for each batch using the softmax activation technique.

### 3.4 Performance Measure

Several metrics are used to evaluate the images, including the precision score, recall, f1 score, roc, auc, and Jaccard score. Figure 6 shows that 13 classes were successfully predicted, with no incorrect predictions and remaining 2 classes got one prediction wrong. As a result, we can see in Table 2 that all of the stated numbers for 13 categories are 100% and differs for remaining two categories. For all leaf species, the average values of the Matthews coefficient and Cohen's kappa coefficient were determined, and both coefficients were 99.69% and 99.68% respectively.

### 3.4.1 Evaluation Metrics

When a set of predicted labels for a sample is compared to the corresponding set of actual labels, the Jaccard similarity coefficient is utilized. It is calculated by dividing the intersection size by the union size of two label sets. Equation (2) gives the Jaccard coefficient.

$$J(T,S) = \frac{|TP|}{|TP|+|FN|+|FP|} \tag{2}$$

$$Sensitivity = \frac{|TP|}{|TP|+|FN|} \tag{3}$$

$$Precision = \frac{|TP|}{|TP|+|FP|} \tag{4}$$

$$F1 = \frac{2\times precision\times recall}{precision+recall} \tag{5}$$

$$Specificity = \frac{|TP|}{|TP|+|FN|} \tag{6}$$

$$Accuracy = \frac{TP+TN}{TP+TN+FP+FN} \tag{7}$$

Where, T, S = Label sets, TP=True Positive, TN=True Negative, FP=False Positive, FN= alse Negative

In Figure 9, a bar chart displaying two more coefficients, the first of which is the Cohen coefficient. Cohen's Kappa (k) is a little-known measure. Measures such as recall, precision and accuracy may not always render a clear view of our model's performance in multi-class classification problems. Another issue that we may encounter is that of imbalanced classes. For example, suppose there are two categories, X and Y, and X only comes up 5% of the time. Because accuracy can be deceiving, we rely on measurements like recall and precision. And we can combine these measurements for single measure called F1_score, but the F1_score lacks an intelligible explanation beyond being the harmonic mean of precision and recall.

The Cohen's kappa statistic is given using Equation (8), is an excellent metric for dealing with both imbalanced class and multi-class issues quite well.

$$k = \frac{p_o-p_e}{1-p_e} \tag{8}$$

Where, the observed agreement is $p_o$, and $p_e$ is an expected agreement. It gives the information about the classifier performance. Cohen's kappa is typically one or less than one, but cannot be more than one. The classifier is ineffective if it returns a value of 0 or less. Its values cannot be interpreted in a uniform fashion.

Second is the Matthews correlation coefficient (MCC) is given using Equation (9). It is a measure of the validity of binary classifications.

$$MCC = \frac{(TP\times TN)-(FP\times FN)}{\sqrt{(TP+FP)(TP+FN)(TN+FP)(TN+FN)}} \tag{9}$$

## 4. RESULTS

Experimental results for the proposed methodology have been given in this section. The confusion matrix for the classification of fifteen classes is shown in Figure 6. A confusion matrix is a table that summarizes how many correct and wrong predictions (or real and anticipated values) a classifier produced. where the diagonal elements represent the proportion of correct classifications for each class. It indicates the percentage of plant species that have been misclassified. It can be seen that 13 classes were predicted correctly and remaining 2 classes 'Betula pubescens', 'Ulmus carpinifolia' misclassified while validation. The results are presented in Table 2.

Table 2: Results on Swedish leaf dataset.

| Type | Precision | Recall | f1-score | Support |
|---|---|---|---|---|
| Acer | 1 | 1 | 1 | 25 |
| Alnus incana | 1 | 1 | 1 | 20 |
| Betula pubescens | 0.952 | 1 | 0.975 | 20 |
| Fagus silvatica | 1 | 1 | 1 | 24 |
| Populus | 1 | 1 | 1 | 24 |
| Populus tremula | 1 | 1 | 1 | 25 |
| Quercus | 1 | 1 | 1 | 27 |
| Salix alba | 1 | 1 | 1 | 21 |
| Salix aurita | 1 | 1 | 1 | 17 |
| Salix sinerea | 1 | 1 | 1 | 28 |
| Sorbus aucuparia | 1 | 1 | 1 | 25 |
| Sorbus intermedia | 1 | 1 | 1 | 20 |
| Tilia | 1 | 1 | 1 | 21 |
| Ulmus carpinifolia | 1 | 0.941 | 0.969 | 17 |
| Ulmus glabra | 1 | 1 | 1 | 24 |
| Average / total | 0.997 | 0.997 | 0.997 | 338 |

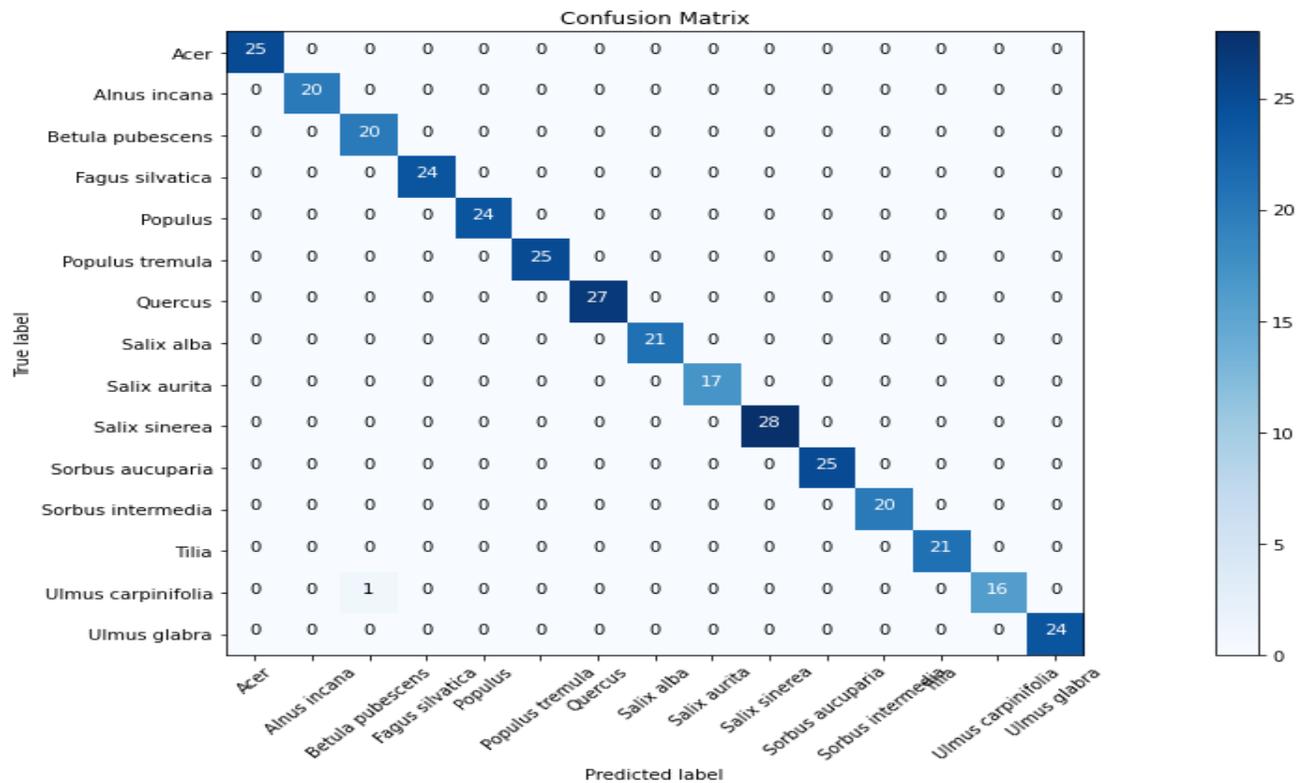

Figure 6. Confusion matrix for the prediction of all leaf classes.

Figure 7 and Figure 8 illustrate the accuracy and loss for each of the training and validation epochs. The accuracy of this model varies initially but later it seems to be constant because values at y-axis of plot range between $50 - 100$ (50% - 100% accuracy) and number of epochs are 50. When we look at the model loss plot, we see the same pattern as accuracy plot. Loss range at y-axis is between $0 - 10$. As the accuracy increases model loss decreases. The model achieves the utmost accuracy possible without over-fitting. The training has been completed at 50 epochs, and the measured classification accuracy is

99.70% at this point. After 15 epochs, the validation and training loss begin to flatten, and when accuracy reaches 99.70%, the loss becomes 0.3.

In Figure 9, we can see bar chart of two more coefficients, first one is cohens coefficient. Cohen's Kappa (k) is an under-utilized, metric.

Further, Bar chart of Jaccard score for all 15 leaf species and ROC-AUC curve for each class vs Rest have been plotted in Figures 10 and 11, respectively.

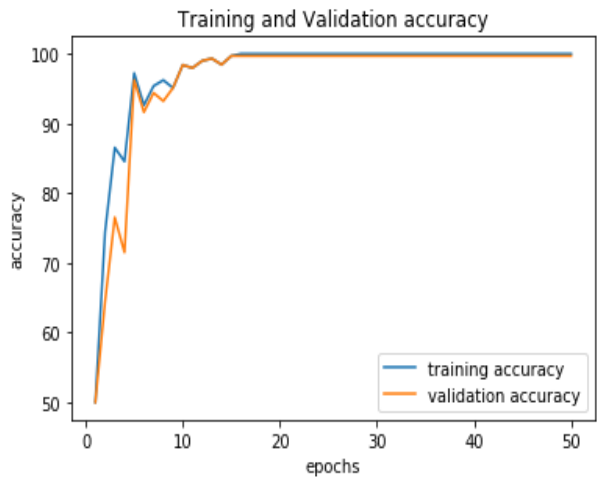

Figure 7. Plot for model accuracy vs epoch.

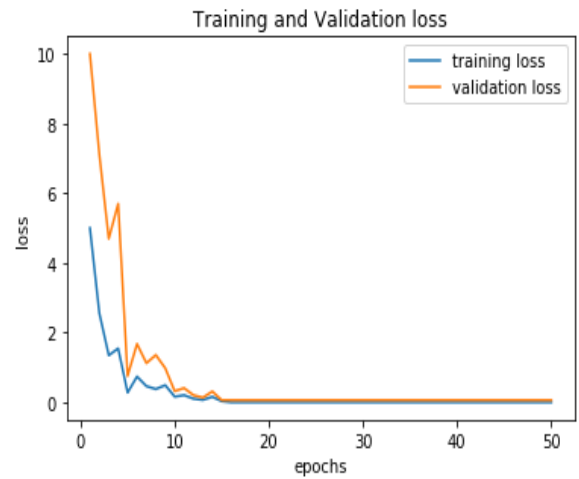

Figure 8. Plot for model loss vs epoch.

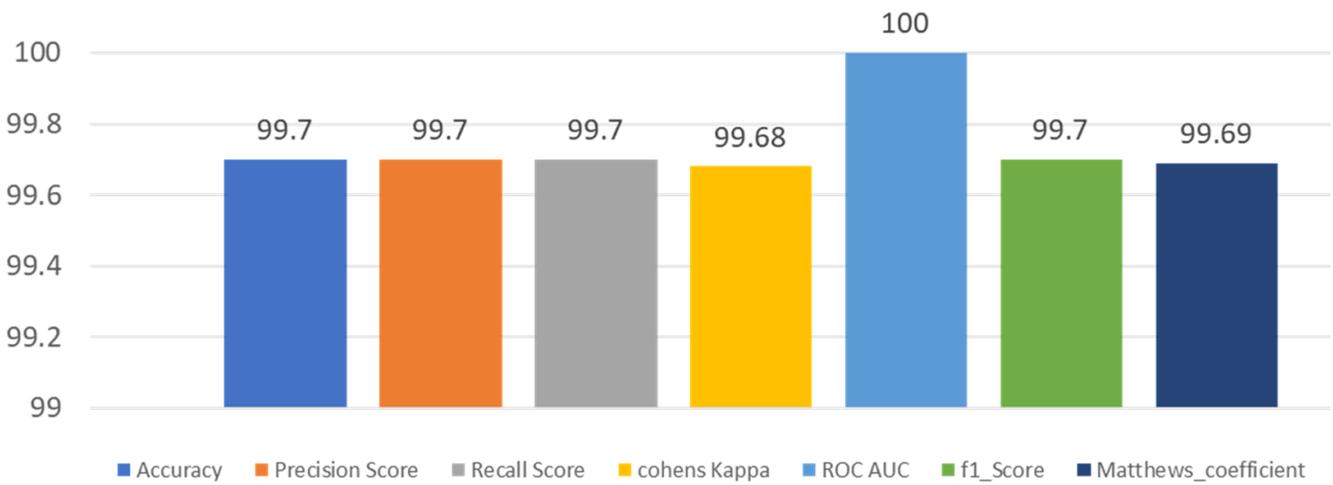

Figure 9. Bar chart of Average of all evaluated values.

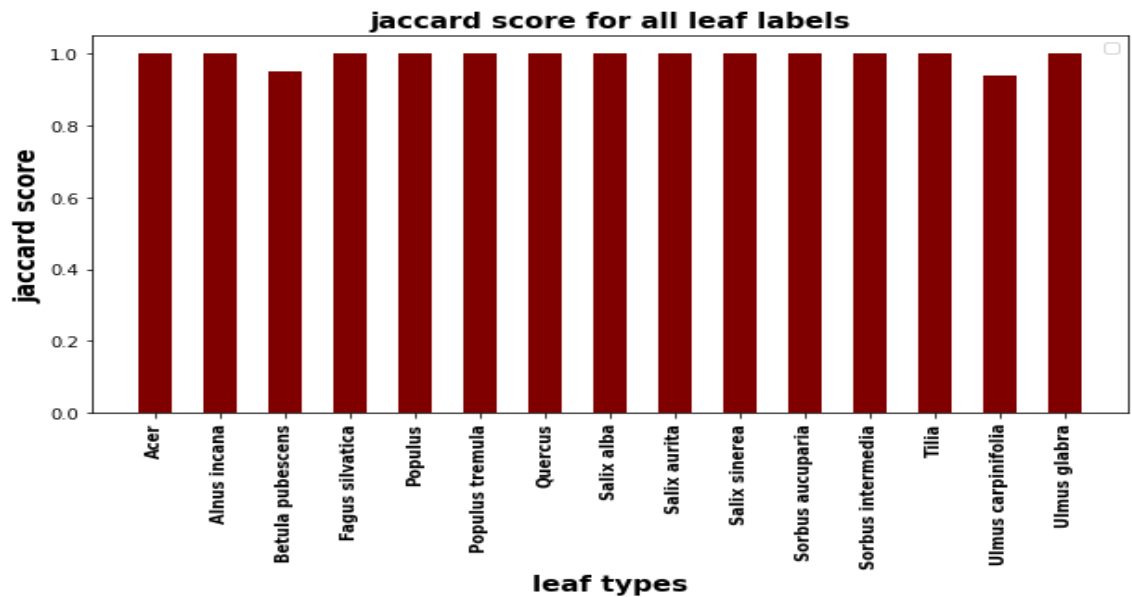

Figure 10. Bar chart of Jaccard score for all 15 leaf species.

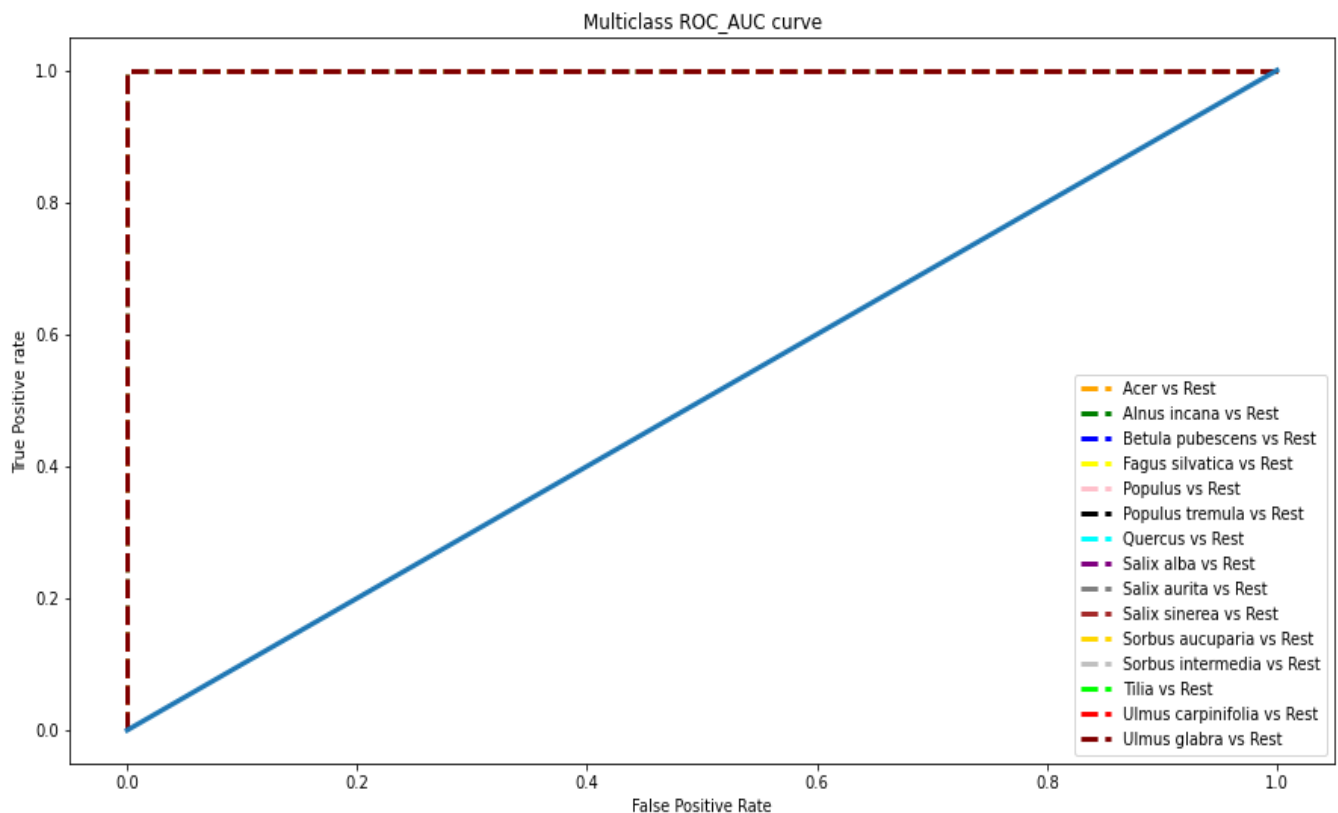

**Figure 11. ROC-AUC curve for each class vs Rest.**

## 5. DISCUSSION

On the Swedish leaf dataset, we tested our method's classification accuracy. 70 percent of the leaves have been used as a training set, while the remaining 30 percent as testing set.

In the Table 3, a comparision of our results with other thirteen existing has been given. We can observe that among all the leaf classification methods, the VGG19 method achieves the highest prediction rate. The prediction rate of the VGG19 model is 6% and 11% higher than the VGG16 [11] and GoogleNet [3] transfer learning methods, respectively. It is 3% and 3.5% higher than the CNN [12] and Dual path CNN [5] methods, respectively. The CNN approaches have roughly the same classification rate as the VGG19 method. because CNN is the best method for features extraction, therefore VGG19 have the better performance than other models because it is also based on DCNN architecture.

As a result, adding other visual parameters including texture, vein, and colour to improve the classification accuracy of the Swedish leaf dataset is required. Our method also outperforms the six deep learning-based leaf recognition models in terms of prediction rate. Our method outperforms the TwoCNN [4] and MSF-CNN [6] methods by roughly 3% and 12%, respectively, in terms of classification rate. We can observe that all of the approaches have a fairly consistent identification rate. As a result, the metric recognition rate demonstrates our method's efficacy in the objective of leaf recognition.

Table 3 demonstrates the performance of the suggested method when compared to other classification methods, including future perceptions. It shows how the suggested model compares to others in terms of accuracy, as well as how existing models can be improved. Our proposed method achieves better performance.

## 6. CONCLUSION

In this work, a plant species identification technique based on transfer learning is proposed and applied to a Swedish leaf dataset. A pre-trained classifier called VGG19 has been used for identifying images of leaves. The work at hand consists of 15 plant species categories and 1125 total leaves, with 75 images from each class. The model uses 70% of the images for training and the remaining 30% for validation. We also inferred from Table 3 and Table 2, that the suggested method has the best accuracy, recall, and precision values when compared to state-of-the-art models. Almost all of the categories are correctly classified with a 99.70% accuracy rate.

Our method's key benefits are its ease of use and great efficiency, making it ideal for large-scale plant species recognition applications. Although the proposed method is not meant to take the role of human taxonomists, it could provide a quick and easy way to identify plants with adequate accuracy. Taxonomists, or people who can accurately identify plant species, are limited, which is impacting research in fields like biotechnology and genetics.

Table 3: Comparison of our method with existing methods.

| Authors | Years | Methods | Accuracy |
|---|---|---|---|
| Lee et al. [4] | 2015 | TwoCNN | 97.54 |
| Jeon and Rhee [3] | 2017 | GoogleNet | 89.5 |
| Shah et al. [5] | 2017 | Dual-path CNN | 96.28 |
| Hu et al. [6] | 2018 | MSF-CNN | 88 |
| Zhu et al. [9] | 2019 | DCNN | 82 |
| Turkoglu and Hanbay [7] | 2019 | LBP | 81.2 |
| Manasa et al. [8] | 2019 | NN | 80 |
| Wang et al. [22] | 2020 | SVM | 98.4 |
| Sumedh et al. [11] | 2021 | VGG16 | 95.75 |
| Dhananjay [12] | 2021 | CNN | 97 |
| Chengzhuan [10] | 2021 | MTD + LBP-HF | 85.74 |
| Kaur [20] | 2021 | SVM | 95 |
| Vidya et al. [21] | 2022 | CNN | 98.04 |
| Our Method | 2022 | VGG-19 | 99.7 |

There are a couple of drawbacks to our approach. It takes a long time to train, and the network architectural weights are pretty huge (in terms of disk/bandwidth). VGG19 is over 575MB in size due to its depth and number of layers. As a result, implementing VGG is a time-consuming process. In the future, we expect to complete all of the tasks associated with our suggested model.

## DISCLOSURE STATEMENT

Author(s) declares that there is no conflict of interest.

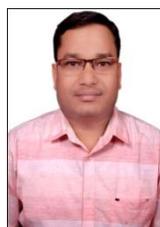

**Bhupendra Singh Kirar** received BE degree in Electronics and Communication Engineering from Samrat Ashok Technological Institute, Vidisha, India in 2004. He received MTech and PhD degree from Maulana Azad National Institute of Technology, Bhopal, India in 2009 and 2019, respectively. He is presently working as a Assistant Professor in the department of Electronics and Communication Engineering, Indian Institute of Information Technology Bhopal, India. He has co-authored a text book on "Basic Electrical & Electronics Engineering", by M/s. BS Publications; Hyderabad in the year 2011. He has published papers in national and international journals under SCI/Scopus. His research interest includes area of image processing and machine learning.

E-mail: bhup17@gmail.com

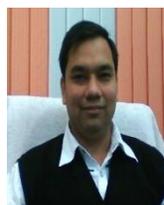

**Dheeraj Kumar Agrawal** received the BE degree with honors in Electronics and Communication Engineering from Rajiv Gandhi Technological University, Bhopal, India in 2001, and MTech and PhD degrees in Electronics and Communication Engineering from Maulana Azad National Institute of Technology, Bhopal, India, in 2005 and 2011, respectively. He is Associate Professor in the department of Electronics and Communication Engineering, Maulana Azad National Institute of Technology, Bhopal, India. He is a member of different professional bodies. He received the Young Scientist award of Madhya Pradesh Council of Science and Technology in the Discipline Engineering Science and Technology in 2011. He served as investigator for different projects undertaken. He has over 18 years of teaching experience in the field of Electronics and Communication Engineering apart from this he has published many papers in national and international journals under SCI/Scopus. His research interests are the areas of image processing and signal processing.

E-mail: profdheerajagrawal@gmail.com

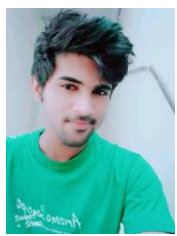

**Thiru Siddharth** pursuing BTech degree from the Department of Computer Science and Engineering, Indian Institute of Information Technology, Bhopal, India. He has experience in industry as well as research. His areas of interest are artificial intelligence, computer science and bioinformatics.

E-mail: thirusid789@gmail.com